\documentclass[11pt]{article}

\usepackage[final]{acl}

\usepackage{times}
\usepackage{latexsym}

\usepackage[T1]{fontenc}

\usepackage[utf8]{inputenc}

\usepackage{microtype}

\usepackage{inconsolata}

\usepackage{graphicx}
\usepackage{amsmath}
\usepackage{hyperref}
\usepackage{url}
\usepackage{multirow}
\usepackage{booktabs}
\usepackage{wrapfig}
\usepackage[table]{xcolor}
\usepackage{lipsum} 

\usepackage{array}
\usepackage{colortbl}
\usepackage{listings}
\definecolor{lightgray}{RGB}{240,240,240}
\definecolor{highlightbg}{RGB}{255,248,220}
\definecolor{codefail}{RGB}{255,235,235}
\definecolor{codepass}{RGB}{235,248,235}
\definecolor{failred}{RGB}{255,235,235}
\definecolor{passgreen}{RGB}{235,248,235}
\definecolor{failheader}{RGB}{200,60,60}
\definecolor{passheader}{RGB}{50,140,80}
\definecolor{envgray}{RGB}{245,245,245}
\definecolor{codebg}{RGB}{250,250,250}
\lstdefinestyle{plaintable}{
  basicstyle=\ttfamily\footnotesize,
  backgroundcolor=\color{white},
  frame=none,
  breaklines=true,
  keepspaces=true,
  columns=fullflexible,
}

\lstdefinestyle{colortable}{
  basicstyle=\ttfamily\scriptsize,
  backgroundcolor=\color{codebg},
  frame=none,
  breaklines=true,
  keepspaces=true,
  columns=fullflexible,
  xleftmargin=4pt,
  xrightmargin=4pt,
}

\usepackage{tabularx}
\newcommand{\Env}[1]{%
  \newline\colorbox{envgray}{\parbox{\dimexpr\linewidth-2\fboxsep\relax}%
  {\textbf{Environment:} \texttt{\scriptsize #1}}}\newline}

\title{RefCon: Iterative Refinement and Contrastive Memory Extraction for Context-Evolving Agent}

\author{
 \textbf{Ubaidillah Ariq Prathama\textsuperscript{1}},
 \textbf{Bo Liu\textsuperscript{1}},
 \textbf{Yeo Boon Hong\textsuperscript{1}},
 \\
 \textbf{Yu-Xuan Huang\textsuperscript{1}},
 \textbf{Yangkai Ding\textsuperscript{1}},
 \textbf{Tao Yu\textsuperscript{1}}
\\
 \textsuperscript{1}Huawei Technologies, Co., Ltd.
\\
}

\begin{document}
\maketitle
\begin{abstract}
Long-horizon agent interactions generate useful but noisy experience, and retraining models to absorb it is expensive. Context-evolving agents therefore need memory extraction methods that improve with more test-time compute without relying on gold labels. We propose RefCon, which combines sequential self-refinement with parallel self-contrast to extract higher-quality memories without gold labels. Evaluated on AppWorld and BFCL-V3 across multiple context-evolving agent frameworks, RefCon delivers strong and consistent gains, including relative improvements of 21.6\% on ACE and 16.6\% on ReMe over no-scaling baselines, while a diversity-focused variant (DivCon) achieves a 35.5\% gain on ReasoningBank. RefCon consistently outperforms existing baselines without ground-truth labels, and generalizes across model scales and to software engineering tasks, where it surpasses even ground-truth baselines. We further analyze the accuracy-token trade-off and scaling behavior, showing RefCon maintains favorable efficiency and continues to improve as more trajectories are used, unlike diversity-only scaling which saturates earlier.
\end{abstract}

\section{Introduction}

\begin{figure*}[t!]
  \centering
  \includegraphics[width=0.95\textwidth]{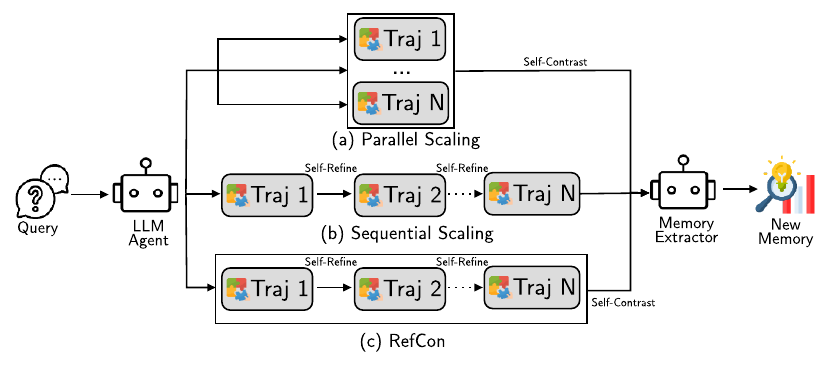}
  \caption{Memory extraction with (a) parallel scaling using self-contrast, and (b) sequential scaling with self-refine, compared to our proposed method (c) RefCon, combining both strength of parallel and sequential scaling.}
  \label{fig:refcon-overview}
\end{figure*}

A few years ago, large language models (LLMs) began to be deployed as agents capable of solving complex tasks through iterative reasoning and tool use. Applications such as web browsing \citep{wei2025browsecomp}, interactive tool calling \citep{patil2025bfcl}, and deep research \citep{phan2025hle} require agents to generate long trajectories of actions and observations. However, these trajectories are typically discarded after task completion, despite often containing reusable insights that could benefit future tasks \citep{shinn2023reflexion,zhao2024expel}. While one possible solution is to incorporate such knowledge through model retraining, updating large language models remains computationally expensive due to their growing scale \citep{hoffmann2022compute}. This motivates lightweight mechanisms that allow agents to accumulate and reuse experience directly from past interactions without costly retraining.

More recently, self-evolving agent approaches integrate knowledge updates directly into the inference loop. In this paradigm, agents continuously write, refine, and retrieve memories to compose an up-to-date context for each new query, enabling continual learning without expensive model retraining \citep{park2023generative,wang2024voyager}. These context-evolving agents typically involve three components: memory extraction, memory retrieval, and memory management \citep{zhang2025ACE,cao2025ReMe, yan202GAM, zhang2025GMemory}. Memory extraction distills raw trajectories into reusable heuristics; memory retrieval selects relevant experiences for the current task; and memory management maintains a compact memory store by filtering redundancy and irrelevant content.
Recent work further explores memory-aware test-time scaling (MaTTS) to improve memory extraction by generating multiple trajectories and selecting higher-quality experiences \citep{ouyang2025ReasoningBank}. This direction builds on broader findings that test-time scaling can improve agent performance \citep{snell2025tts,zhu2025tts}.

Despite recent progress, several challenges remain. Some methods rely on gold labels or ground-truth signals to extract memories \citep{cao2025ReMe,zhang2025ACE,cai2025FLEX}, which are often unavailable in real-world settings, while replacing them with LLM-as-a-Judge \citep{zheng2023judging} can be unreliable for long-horizon interactions where trajectories appear plausible yet fail to achieve the intended goal \citep{ouyang2025ReasoningBank,yang2025MUSE,liu2025SMITH}.
ReasoningBank \citep{ouyang2025ReasoningBank} addresses this by introducing self-contrast reasoning over multiple trajectories for parallel scaling (Figure~\ref{fig:refcon-overview} (a)) and self-refinement for sequential scaling (Figure~\ref{fig:refcon-overview} (b)). However, while prior work explores both paradigms \citep{stein2025EGuR,wu2025EvolveR,wang2025SAGE,yan202GAM,tang2025AgentKB}, they are typically studied in isolation despite being complementary: parallel scaling improves trajectory diversity while sequential scaling enhances trajectory quality. Evaluating memory extraction effectively requires benchmarks with two key properties: (1) task recurrence, so memories extracted from earlier tasks transfer to later ones, and (2) trajectory ambiguity, where single trajectories may contain misleading signals that contrastive multi-trajectory analysis must disambiguate. These properties enable distinguishing between good and poor extraction methods.

To address these limitations, we propose RefCon, an extension of memory-aware test-time scaling (MaTTS) for memory extraction that combines the benefits of sequential and parallel scaling. RefCon first applies sequential scaling to generate 
$N$ trajectories using self-refine \citep{madaan2023refine}, as illustrated in Figure~\ref{fig:refcon-overview}. These trajectories are then jointly analyzed through parallel scaling, where the memory extractor performs self-contrast reasoning \citep{chen2020contrast} to identify improvements across trajectories and distill reusable insights. This design encourages both trajectory diversity and quality, enabling more reliable memory extraction.
We further introduce a variant, DivCon, which promotes diversity by instructing the agent to explore alternative strategies instead of refining previous trajectories. For memory retrieval, we adopt a retrieve–rerank–rewrite pipeline that improves relevance by reranking retrieved memories and rewriting them into a concise form. Overall, our contributions are twofold: 
\begin{itemize}
    \item  We propose \textbf{RefCon}, a memory-aware test-time scaling framework that unifies parallel and sequential scaling to improve memory extraction quality. We will release it as a plug-in for agentic systems to support context-evolving methods, enabling easier adoption and future research.
    \item Extensive experiments demonstrate that RefCon consistently outperforms existing baselines in the without-ground-truth setting. Compared with the no-scaling setting, RefCon improves performance by 21.6\% on ACE and 16.6\% on ReMe, while the variant DivCon achieves a 35.5\% gain on ReasoningBank. 

\end{itemize}

\section{Related Works}

\textbf{Memory Extraction.} Modern memory extraction has transitioned from passive logging to active insight distillation. ACE \citep{zhang2025ACE} decouples reflection from curation to synthesize memory updates, while ReasoningBank \citep{ouyang2025ReasoningBank} highlights the importance of integrating cautionary insights from failed trajectories. Other frameworks distill insights at varying granularities: MUSE \citep{yang2025MUSE} and FLEX \citep{cai2025FLEX} separate high-level heuristics from execution details, EGuR \citep{stein2025EGuR} learns task-specific workflows, and SAGE \citep{wang2025SAGE} and SMITH \citep{liu2025SMITH} generate reusable code snippets. ReasoningBank \citep{ouyang2025ReasoningBank} also introduces memory-aware test-time scaling (MaTTS), utilizing self-contrast reasoning \citep{chen2020contrast} for parallel scaling and self-refine \citep{madaan2023refine} for sequential scaling, a foundation adopted and extended by many follow-up works in both parallel \citep{stein2025EGuR,wu2025EvolveR,cao2025ReMe,liu2025SMITH,cai2025tfGRPO,wang2025SAGE} and sequential \citep{cao2025ReMe,yang2025MUSE,yan202GAM,tang2025AgentKB} settings.

\textbf{Memory Retrieval.} Earlier approaches either inserted all memories into the prompt \citep{suzgun2025DynamicCheatsheet,zhang2025ACE} or relied on basic embedding-similarity retrieval \citep{ouyang2025ReasoningBank}. Since then, retrieval has evolved into dynamic and hierarchical architectures: ReMe \citep{cao2025ReMe} applies a retrieve–rerank–rewrite pipeline, GAM \citep{yan202GAM} employs "Researcher" agents for deep dives into historical data, G-Memory \citep{zhang2025GMemory} leverages graph traversals, AgentKB \citep{tang2025AgentKB} separates workflow patterns from execution details, and EvolveR \citep{wu2025EvolveR} treats retrieval as a learned tool-call via GRPO.

\textbf{Memory Management.} To prevent unbounded memory growth, recent systems treat memory as a self-optimizing component. ACE \citep{zhang2025ACE} and FLEX \citep{cai2025FLEX} delegate insert, update, and delete operations to a dedicated agent, while ReMe \citep{cao2025ReMe} and EvolveR \citep{wu2025EvolveR} prune memories based on retrieval utility scores. Despite its importance, explicit memory management remains underexplored in many current systems.

\section{Methodology}

\begin{figure*}[t]
  \centering
  \includegraphics[width=0.85\textwidth]{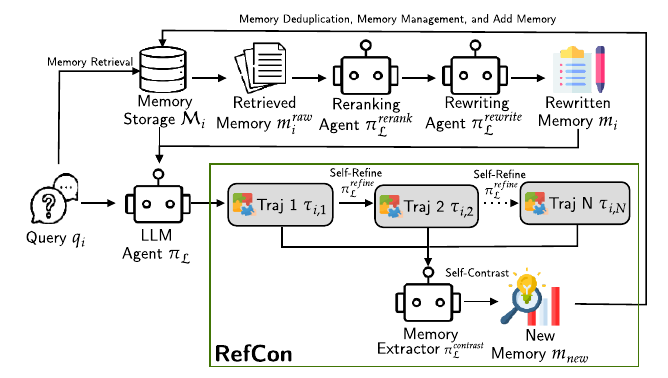}
  \caption{Overview of context-evolving agent workflow utilizing RefCon scaling for memory extraction.}
  \label{fig:refcon-workflow}
\end{figure*}

\subsection{Problem Formulation}

In this work, we focus on the test-time learning (TTL) paradigm \citep{wu2024ttl,wang2025ttl}, where a model continues to improve its performance during inference without access to ground-truth supervision. Specifically, we consider a setting in which an agent must adapt to a continuous stream of task queries $Q = \{q_1, q_2, \dots, q_N\}$ encountered sequentially in an online environment. Unlike traditional supervised learning, this setting precludes access to future queries or ground-truth labels. Instead, the agent must \textbf{self-evolve} by critically analyzing its past trajectories and leveraging self-verification mechanisms. The primary objective is to enable the system to extract and retain useful insights from prior interactions, thereby avoiding redundant rediscovery of successful strategies or the repetition of past failures. 

Specifically, we instantiate an LLM-based agent $\pi_{\mathcal{L}}$ that solves each query through iterative tool use and reasoning in a ReAct loop, while reusing accumulated experience across tasks. To model this cross-task learning process, we define a memory state initialized as $\mathcal{M}_0 = \emptyset$. At task $i$, the agent conditions on the current memory $\mathcal{M}_i$ (appended to the prompt) to produce a trajectory for the query. After the task is completed, the trajectory is distilled into memory updates, yielding the next state $\mathcal{M}_{i+1}$. Thus, memory evolves at the task level as $\mathcal{M}_0, \mathcal{M}_1, \ldots$, and the agent for task $i$ is parameterized by $\pi_{\mathcal{L}}(\cdot; \mathcal{M}_i, \mathcal{A})$, where $\mathcal{A}$ is the available action/tool space.

\subsection{RefCon for Memory Extraction}
We propose an iterative \textbf{ref}inement and \textbf{con}trastive (RefCon) framework, which generates trajectories for our memory extraction process as shown in Figure~\ref{fig:refcon-workflow} green zone.
The memory extraction process within the test-time learning paradigm is formulated as an iterative refinement procedure that generates a sequence of trajectories. For a given task query $q_i$, the agent first produces an initial trajectory by conditioning its policy $\pi_{\mathcal{L}}$ on the rewritten memory $m_i$, which is retrieved from the current memory state $\mathcal{M}_i$:
\begin{equation}
\tau_{i,1} = \pi_{\mathcal{L}}(q_i, m_i).
\end{equation}

Subsequently, the agent iteratively refines its previous trajectory to generate improved problem-solving attempts:
\begin{equation}
\tau_{i,n} = \pi_{\mathcal{L}}^{\text{refine}}(q_i, m_i, \tau_{i,n-1}), \quad n \in \{2, \dots, N\},
\end{equation}
where $\pi_{\mathcal{L}}^{\text{refine}}$ represents the self-refinement policy that evaluates and improves upon the preceding trajectory. The resulting trajectory set $\{\tau_{i,1}, \tau_{i,2}, \dots, \tau_{i,N}\}$
captures a progression of reasoning attempts, evolving from initial exploration to increasingly refined and self-corrected strategies.

Once the $N$ trajectories are generated, a self-contrast reasoning mechanism is applied to distill high-value insights for memory extraction. Specifically, the agent's memory extractor compares the differences between \emph{successful} and \emph{unsuccessful} steps across the $N$ trajectories, identifying key turning points or reasoning patterns that lead to successful outcomes. Through this contrastive analysis, the agent summarizes these insights into a set of candidate memories $m_{new}$, formally defined as
\begin{equation}
m_{new} = \pi_{\mathcal{L}}^{\text{contrast}}(\tau_{i,1}, \dots, \tau_{i,N}),
\end{equation}
where $\pi_{\mathcal{L}}^{\text{contrast}}(\cdot)$ denotes the contrastive memory extraction operator that distills actionable knowledge from multiple trajectories.

Before finalizing the knowledge state $\mathcal{M}_{i+1}$, the system performs a memory deduplication step using embedding-based similarity ($\text{Sim}$) with a threshold $\epsilon$ to remove redundant memories. Specifically, the candidate memories $m_{new}$ are compared against both the existing memory $\mathcal{M}_i$ (inter-memory comparison) and among themselves (intra-memory comparison). Only non-redundant memories are retained, ensuring that memory remains concise and informative. The update rule is defined as:
\begin{equation}
\begin{split}
\mathcal{M}_{i+1} &= \mathcal{M}_i \cup \{m \in m_{\text{new}} \mid \\
&\quad \text{Sim}(m, \mathcal{M}_i \cup m_{\text{new}} \setminus \{m\}) < \epsilon\}.
\end{split}
\end{equation}
The resulting memory $\mathcal{M}_{i+1}$ now contains memories from all previous $i+1$ task. Each memory entry contains two key components: (i) \textit{when to use}, a short explanation of when to use the memory, also used as an index; (ii) \textit{content}, describing the insight from previous trajectory to guide agent in similar situation. These memories also contains other metadata such as frequency and utility.

\textbf{DivCon}. We also propose an alternative trajectory generation strategy named \textbf{DivCon}. Unlike self-refine in RefCon, DivCon uses self-diversity to enhance memory extraction through self-contrasting highly diverse trajectories. The difference with RefCon relies on Eq. 2, where DivCon prompts the agent to explore other solutions:
\begin{equation}
\tau_{i,n} = \pi_{\mathcal{L}}^{\text{diversity}}(q_i, m_i, \tau_{i,n-1}).
\end{equation}

\subsection{Memory Retrieval and Management}
The retrieval process serves as the initial phase of the test-time learning cycle, augmenting the agent's policy $\pi_{\mathcal{L}}$ with the most relevant historical insights before solving a new task. We adopt a \textit{retrieve–rerank–rewrite} strategy. Given a query $q_i$, the system first retrieves candidate memories from the current memory state $\mathcal{M}_i$ using embedding similarity. A reranking agent then evaluates these candidates and selects the most relevant insights according to the policy’s internal ranking logic. Finally, a rewriting agent consolidates the reranked memories to produce a concise and query-aligned memory:
\begin{equation}
m_i = \pi_{\mathcal{L}}^{\text{rewrite}}(\pi_{\mathcal{L}}^{\text{rerank}}(\text{Sim}(\mathcal{M}_i, q_i), q_i), q_i).
\end{equation}
The resulting memory $m_i$ summarizes the most relevant prior experience and is used to guide the agent when solving the current task.
To maintain a high-quality knowledge base, we implement a utility-based deletion strategy to prune ineffective memories. This ensures that the agent's behavior is shaped only by high-utility memories. We define the utility of an experience $E \in \mathcal{M}$ based on two metrics: the total number of retrievals $f(E)$ and the historical utility $u(E)$, which increments by 1 whenever its recall contributes to a successful task completion. An experience is flagged for removal if its average utility falls below a predefined threshold $\beta$, provided it has been retrieved at least $\alpha$ times.Using the established notation, the deletion function $\phi_{remove}$ is formulated as:\begin{equation}
\phi_{remove}(E) = \begin{cases} 1 \left[ \frac{u(E)}{f(E)} \leq \beta \right], & \text{if } f(E) \geq \alpha, \\ 0, & \text{otherwise.} \end{cases}
\end{equation}

\section{Experiments}

\begin{table*}[!t]
\centering
\footnotesize
\caption{Performance of context-evolving agent methods on AppWorld and BFCL-V3 Using GLM-4.6.}
\label{tab:main-results}
\begin{tabular}{l cccc cc}
\toprule
\multirow{3}{*}{\textbf{Methods}} & \multicolumn{4}{c}{\textbf{AppWorld}} & \multicolumn{2}{c}{\textbf{BFCL-V3}} \\
\cmidrule(lr){2-5} \cmidrule(lr){6-7}
& \multicolumn{2}{c}{Avg@3} & \multicolumn{2}{c}{Pass@3} & Avg@3 & Pass@3 \\
\cmidrule(lr){2-3} \cmidrule(lr){4-5}
& TGC & SGC & TGC & SGC &  &  \\
\midrule
\rowcolor{gray!15}\multicolumn{7}{c}{Baseline} \\
ReAct & $63.77^{\pm 4.60}$ & $35.10^{\pm 10.80}$ & 87.72 & 68.42 & $52.67^{\pm 4.11}$ & 62.00 \\
\midrule
\rowcolor{gray!15}\multicolumn{7}{c}{With Gold Labels/Ground-Truth} \\
ACE (with Ground-Truth) & $\textbf{78.00}^{\pm \textbf{5.34}}$ & $50.9^{\pm 8.96}$ & \textbf{96.49} & \textbf{89.47} & $66.67^{\pm 4.71}$ & \textbf{82.00} \\
ReMe (Sequential Scaling) & $75.43^{\pm 10.81}$ & $\textbf{57.93}^{\pm \textbf{14.90}}$ & \textbf{96.49} & \textbf{89.47} & $\textbf{67.33}^{\pm \textbf{0.94}}$ & \textbf{82.00} \\
ReMe (Parallel Scaling) & $62.00^{\pm 0.85}$ & $36.80^{\pm 0.00}$ & 84.21 & 68.42 & $58.67^{\pm 1.89}$ & 68.00 \\
\midrule
\rowcolor{gray!15}\multicolumn{7}{c}{Without Gold Labels/Ground-Truth} \\
ACE (without Ground-Truth) & $59.63^{\pm 2.9}$ & $42.10^{\pm 4.33}$ & 77.19 & 73.68 & $58.00^{\pm 5.89}$ & 80.00 \\
ReasoningBank & $56.13^{\pm 1.43}$ & $42.10^{\pm 4.33}$ & 80.70 & 68.42 & $60.00^{\pm 7.12}$ & \textbf{82.00} \\
ReasoningBank (Sequential Scaling) & $69.03^{\pm 1.65}$ & $43.87^{\pm 2.50}$ & 84.21 & 73.68 & $64.00^{\pm 1.63}$ & 74.00 \\
ReasoningBank (Parallel Scaling) & $67.27^{\pm 2.20}$ & $45.63^{\pm 10.81}$ & 85.96 & \textbf{78.95} & $\textbf{65.33}^{\pm \textbf{4.11}}$ & 72.00 \\
DivCon & $71.93^{\pm 5.76}$ & $47.37^{\pm 4.29}$ & \textbf{92.98} & \textbf{78.95} & $58.67^{\pm 3.40}$ & 76.00 \\
RefCon & $\textbf{74.30}^{\pm \textbf{0.94}}$ & $\textbf{52.67}^{\pm \textbf{7.45}}$ & \textbf{92.98} & \textbf{78.95} & $63.33^{\pm 0.94}$ & \textbf{82.00} \\
\bottomrule
\end{tabular}
\end{table*}

\subsection{Setup}


\textbf{Datasets.} \textbf{Datasets.}  Evaluating memory extraction quality requires benchmarks with two properties: \textbf{recurrence} (structurally similar tasks across the sequence so extracted memories are reusable) and \textbf{trajectory ambiguity} (tasks complex enough that single trajectories may contain misleading signals), making contrastive multi-trajectory analysis necessary. We use two datasets emphasizing multi-step agent interaction: AppWorld \citep{trivedi2024appworld}, which contains app-like tasks requiring tool use across multiple steps, and BFCL-V3 \citep{patil2025bfcl}, which focuses on tool-calling and decision sequences. We use the AppWorld dev split and sample 50 tasks from the BFCL-V3 multi-turn travel subset, which emphasizes recurrence. We report Avg@3 and Pass@3 on BFCL-V3, and Avg@3, Pass@3, Task Goal Completion (TGC) and Scenario Goal Completion (SGC) on AppWorld, with standard deviation across three runs.  AppWorld SGC captures recurrence as each scenario contains similar tasks, and further exhibits trajectory ambiguity where similar trajectories may converge on conflicting answers (Table~\ref{tab:trajectory-diverging}, Appendix~\ref{sec:appendix-improvement}). We additionally ablate on SWE-bench Verified (Mini) \citep{jimenez2024swebench, hobbhahn2024SWEBenchMini} to evaluate generalization under weaker feedback and higher trajectory ambiguity.

\textbf{Models.} We use GLM-4.6 \citep{zai2024glm46} as the backbone LLM for agent reasoning, tool use, and memory extraction, and OpenAI text-embedding-3-small \citep{openai2024textembedding3small} for retrieval indexing and similarity search. GLM-4.6 provides strong instruction following and multi-step reasoning, while text-embedding-3-small offers high-quality semantic representations for effective memory retrieval. We also experiment using Gemma 4 31B \citep{googlegemma4} and Qwen3.5 9B \citep{qwen35blog} for generalization to medium and small model.

\textbf{Baselines \& Hyperparameters.} We use ReAct as base and evaluate against three context-evolving agents: ACE \citep{zhang2025ACE}, which extracts memories via a reflector–curator pipeline; ReasoningBank \citep{ouyang2025ReasoningBank}, which introduces MaTTS with embedding-based retrieval; and ReMe \citep{cao2025ReMe}, which further improves retrieval via a retrieve–rerank–rewrite pipeline. We employ a scaling factor of 3 (except 2 for DivCon). We follow the original implementations for results in Table~\ref{tab:main-results}. For Table~\ref{tab:matts-appworld}, we disable ground-truth usage and integrate MaTTS for memory extraction. Further details on baseline methods, hyperparameters, and prompt templates are provided in Appendices~\ref{sec:appendix-example},~\ref{sec:appendix-hyperparams}, and~\ref{sec:appendix-prompt}.


\begin{table*}[t]
\centering
\small
\caption{Comparison of RefCon and DivCon against MaTTS methods for context-evolving agents on AppWorld.}
\label{tab:matts-appworld}
\begin{tabular}{l cccc}
\toprule
\multirow{3}{*}{\textbf{Methods}} & \multicolumn{4}{c}{\textbf{AppWorld}} \\
\cmidrule(lr){2-5}
& \multicolumn{2}{c}{Avg@3} & \multicolumn{2}{c}{Pass@3} \\
\cmidrule(lr){2-3} \cmidrule(lr){4-5} 
& TGC & SGC & TGC & SGC \\
\midrule
\rowcolor{gray!15}\multicolumn{5}{c}{ACE} \\
ACE (Without Scaling) & $59.63^{\pm 2.9}$ & $42.10^{\pm 4.33}$ & 77.19 & 73.68 \\
ACE (Parallel Scaling) & $63.73^{\pm 9.55}$ & $43.87^{\pm 13.83}$ & 84.21 & \textbf{78.95} \\
ACE (Sequential Scaling) & $64.30^{\pm 1.55}$ & $42.10^{\pm 7.45}$ & 84.21 & \textbf{78.95} \\
ACE (DivCon) & $69.00^{\pm 2.16}$ & $42.10^{\pm 4.33}$ & 91.23 & \textbf{78.95} \\
ACE (RefCon) & $\textbf{72.53}^{\pm \textbf{5.82}}$ & $\textbf{43.87}^{\pm \textbf{6.57}}$ & \textbf{92.98} & \textbf{78.95} \\
\midrule
\rowcolor{gray!15}\multicolumn{5}{c}{ReasoningBank} \\
ReasoningBank (Without Scaling) & $56.13^{\pm 1.43}$ & $42.10^{\pm 4.33}$ & 80.70 & 68.42 \\
ReasoningBank (Parallel Scaling) & $67.27^{\pm 2.20}$ & $45.63^{\pm 10.81}$ & 85.96 & 78.95 \\
ReasoningBank (Sequential Scaling) & $69.03^{\pm 1.65}$ & $43.87^{\pm 2.50}$ & 84.21 & 73.68 \\
ReasoningBank (DivCon) & $\textbf{76.03}^{\pm \textbf{2.21}}$ & $\textbf{52.67}^{\pm \textbf{0.00}}$ & \textbf{94.74} & \textbf{89.47} \\
ReasoningBank (RefCon) & $71.93^{\pm 3.80}$ & $49.13^{\pm 13.13}$ & 92.98 & 84.21 \\
\midrule
\rowcolor{gray!15}\multicolumn{5}{c}{ReMe} \\
ReMe (Without Scaling) & $63.73^{\pm 3.30}$ & $43.87^{\pm 2.50}$ & 85.96 & 73.68 \\
ReMe (Parallel Scaling) & $70.76^{\pm 3.57}$ & $42.10^{\pm 4.33}$ & 91.23 & 84.21 \\
ReMe (Sequential Scaling) & $69.67^{\pm 3.30}$ & $47.37^{\pm 4.29}$ & 91.23 & \textbf{78.95} \\
ReMe (DivCon) & $71.93^{\pm 5.76}$ & $47.37^{\pm 4.29}$ & \textbf{92.98} & \textbf{78.95} \\
ReMe (RefCon) & $\textbf{74.30}^{\pm \textbf{0.94}}$ & $\textbf{52.67}^{\pm \textbf{7.45}}$ & \textbf{92.98} & \textbf{78.95} \\
\bottomrule
\end{tabular}
\end{table*}

\subsection{Main Comparison}

\begin{figure}[t]
  \centering
  \includegraphics[width=\columnwidth]{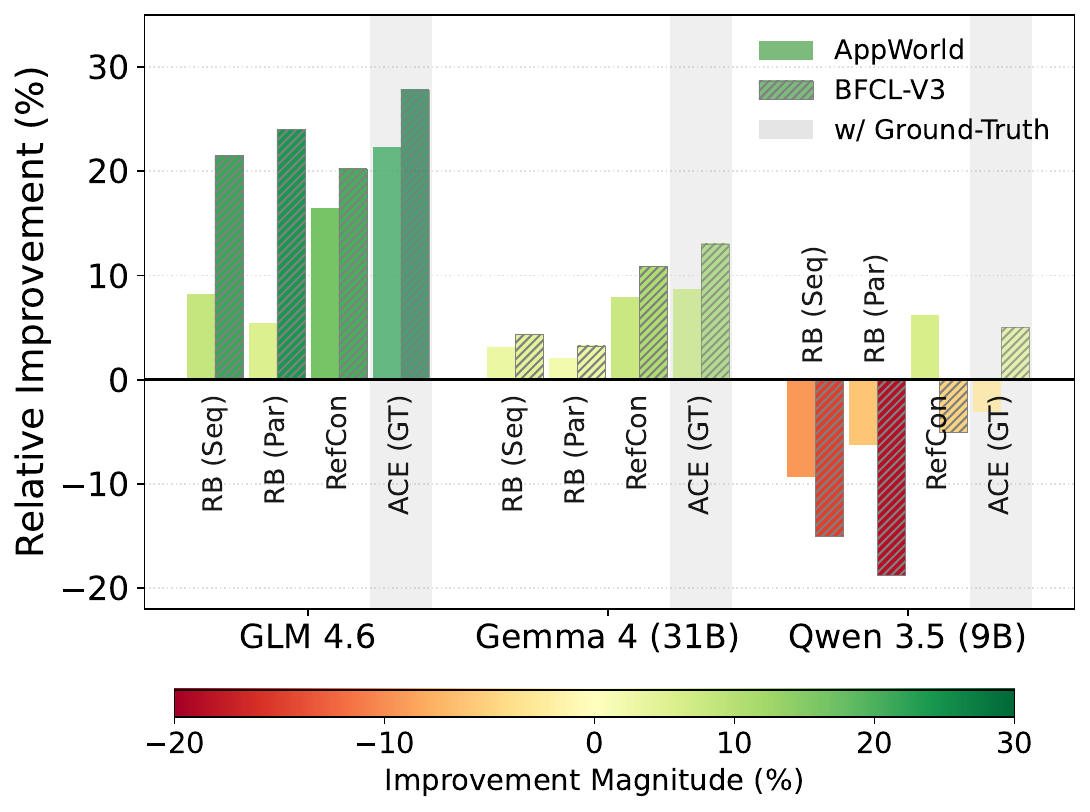}
  \caption{Relative improvement of compared to ReAct baseline across different frameworks and LLMs.}
  \label{fig:main-comparison}
\end{figure}

Table~\ref{tab:main-results} shows that while ground-truth methods remain strong, RefCon achieves the best overall performance without ground-truth labels, surpassing some ground-truth baselines (e.g., ReMe Parallel) and nearly matching the top AppWorld scores. Relative to ReAct, RefCon improves AppWorld Avg@3 by 16.51\% (TGC) and 50.06\% (SGC), and BFCL-V3 Pass@3 by 32.36\%. Against ReasoningBank (Sequential), RefCon improves AppWorld Avg@3 by up to 20.06\% and BFCL-V3 Pass@3 by 13.89\%, though it trails ReasoningBank (Parallel) in BFCL-V3 Avg@3 by 3.06\%, likely because BFCL-V3's narrow distribution favors contrastive diversity over refinement. DivCon consistently ranks second in the without-ground-truth setting except on BFCL-V3. As shown in Figure~\ref{fig:main-comparison}, these trends hold across Gemma 4 31B and Qwen3.5 9B (Tables~\ref{tab:gemma-results}--\ref{tab:qwen-results}, Appendix~\ref{sec:appendix-refcon-divcon}), with RefCon remaining competitive with ACE (ground-truth) across all tested LLMs, though gains diminish on Qwen3.5 9B, suggesting memory extraction quality is contingent on base model capability.

Several anomalies are worth noting. On BFCL-V3, self-refinement is less reliable and can degrade correct trajectories, explaining the underperformance of RefCon, DivCon, and ReasoningBank (Sequential). On AppWorld, ACE and ReasoningBank without scaling fall below ReAct, confirming that MaTTS is critical — single-trajectory extraction without labels risks reinforcing incorrect behavior. Finally, ReMe's parallel scaling underperforms even with ground-truth because it only applies self-contrast when trajectory scores differ, leaving the contrastive signal frequently absent. Additional analysis is provided in Appendix~\ref{sec:appendix-improvement}.

\subsection{RefCon's Combined Scaling Advantage}

In this subsection, we only adopt the memory format and memory retrieval strategy from each method. However, we modify the memory extraction part to adopt parallel and sequential scaling as introduced by \citep{ouyang2025ReasoningBank} and also RefCon and DivCon as introduced by us. We don't use any gold-label or ground-truth to ensure fair comparison in this section. 
We compare the effect of MaTTS methods for each context-evolving agent methods as shown in Table~\ref{tab:matts-appworld} and provide detailed results for each run in Appendix~\ref{sec:appendix-refcon-divcon}.

Across agents, RefCon is the strongest variant on ACE (21.6\% relative improvement over no scaling) and ReMe (16.6\%), while DivCon is the top performer on ReasoningBank (35.5\%), with the runner-up alternating between the two, suggesting complementary strengths: RefCon benefits more when iterative refinement yields progressively better trajectories, whereas DivCon shines when diversity exposes contrasting reasoning patterns. Notably, ACE originally lacks any scaling mechanism, yet adding MaTTS variants substantially improves performance without changing its reflector/curator pipeline, implying gains come from richer trajectory evidence rather than architectural changes and reinforcing that test-time scaling is a broadly effective lever for memory quality across context-evolving agents.

DivCon remains competitive with a smaller scaling factor (2), as contrastive extraction benefits most from genuinely diverse trajectories. Temperature-based parallel sampling can be too homogeneous when the model strongly prefers a single solution. DivCon performs particularly well with ReasoningBank due to its higher-level memory format, which encourages exploring alternative trajectories without contradicting existing memories. We still prioritize RefCon in the main results as DivCon is less stable and exhibits weaker scaling behavior, as will be discussed in the scaling law analysis. We further analyze RefCon's superior ability to resolve recurring problems in Appendix~\ref{sec:appendix-refcon-divcon} (Figure~\ref{fig:task-improvement}), showing that self-contrast memory extraction produces the steepest performance gains between consecutive similar tasks, while sequential scaling fails to maintain consistent improvements due to the limitations of extracting memory from a single trajectory.

\subsection{Generalization to Coding Tasks}

\begin{table}[!t]
\centering
\footnotesize
\caption{Performance of context-evolving agent methods on SWE-Bench-Verified (Mini) using GLM-4.6.}
\label{tab:swe-results}
\begin{tabular}{l ccc}
\toprule
\textbf{Method} & \textbf{Iter 1} & \textbf{Iter 2} & \textbf{Iter 3} \\
\midrule
\rowcolor{gray!15}\multicolumn{4}{c}{Methods Without Refinement} \\
ReAct                    & $50.67^{\scriptstyle \pm 1.15}$ & --- & --- \\
RB (Par) & $54.00^{\scriptstyle \pm 2.00}$ & --- & --- \\
ACE (w/ GT)  & $57.33^{\scriptstyle \pm 1.15}$ & --- & --- \\
\midrule
\rowcolor{gray!15}\multicolumn{4}{c}{Methods With Refinement} \\
Self-Refine                & $50.67^{\scriptstyle \pm 1.15}$ & $52.67^{\scriptstyle \pm 1.15}$ & $54.00^{\scriptstyle \pm 0.00}$ \\
RB (Seq) & $54.00^{\scriptstyle \pm 0.00}$ & $54.00^{\scriptstyle \pm 0.00}$ & $56.67^{\scriptstyle \pm 1.15}$ \\
ReMe (Seq) & $54.67^{\scriptstyle \pm 1.15}$ & $56.67^{\scriptstyle \pm 1.15}$ & $58.67^{\scriptstyle \pm 1.15}$ \\
RefCon                     & $57.33^{\scriptstyle \pm 1.15}$ & $58.00^{\scriptstyle \pm 0.00}$ & $\textbf{60.00}^{\scriptstyle \pm \textbf{0.00}}$ \\
\bottomrule
\end{tabular}
\end{table}

We evaluate RefCon's generalizability on SWE-bench Verified (Mini), a challenging benchmark where feedback is sparse and trajectory correctness is difficult to assess without executing test cases. Although recurrence is less explicit than in AppWorld, software engineering tasks still involve underlying routines where memory is useful. Using GLM 4.6 with mini-SWE-agent \citep{yang2024sweagent}, Table~\ref{tab:swe-results} shows RefCon outperforms all baselines including ACE and ReMe (Sequential) which use ground-truth. We also include vanilla self-refine as a baseline to isolate gains from context-evolving methodology versus standard refinement alone.

While vanilla self-refine shows consistent but slow gains with a low initial success rate matching ReAct, ReasoningBank (Sequential) and RefCon do not see an immediate increase in the second iteration. Instead, they optimize trajectory efficiency during this stage, which subsequently facilitates higher success rates by the third iteration. RefCon's strong first-iteration performance validates its memory extraction and retrieval mechanisms. Its self-contrast reasoning across multiple trajectories enables it to outperform ACE even when ACE is provided ground-truth code patches.

\subsection{RefCon Accuracy-Token Tradeoff}

\begin{figure*}[t]
  \centering
  \includegraphics[width=\textwidth]{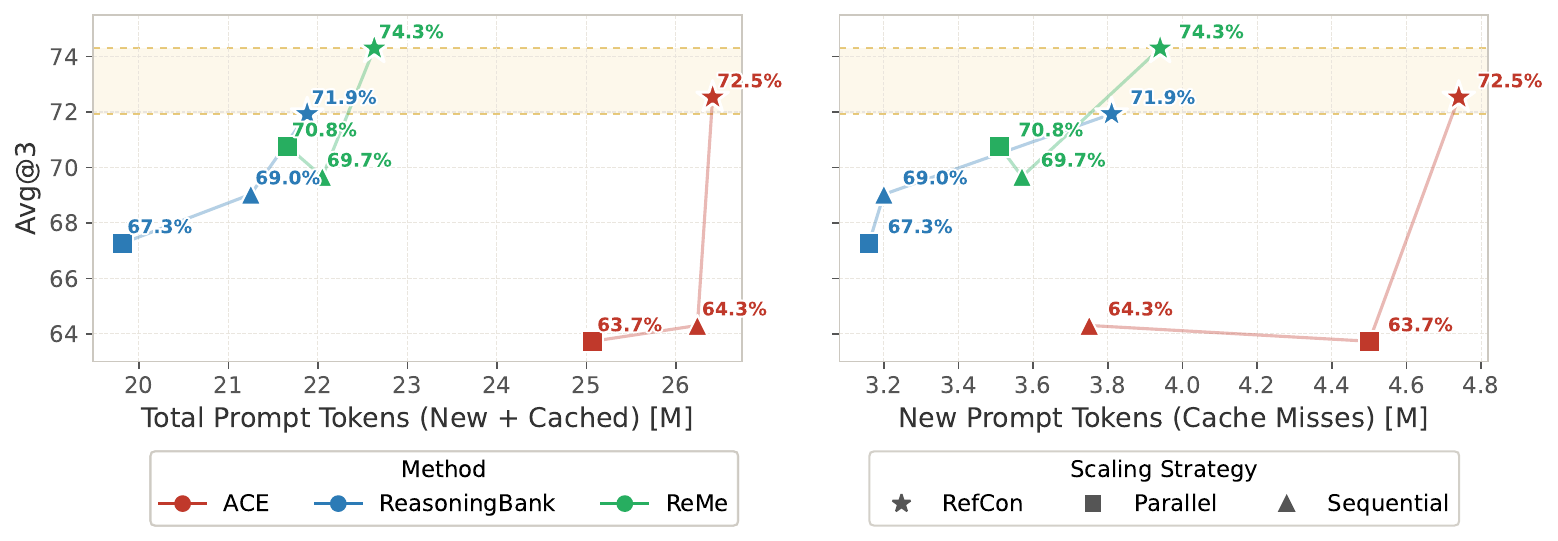}
  \caption{Accuracy (Avg@3) and prompt token tradeoff for each MaTTS methods.}
  \label{fig:acc-token-tradeoff}
\end{figure*}

As shown in Figure~\ref{fig:acc-token-tradeoff}, RefCon consistently achieves the highest accuracy across all methods (72.53\% on ACE, 71.93\% on ReasoningBank, 74.30\% on ReMe) with minimal token overhead of only 0.17M--1.18M ($\le$3\%) over sequential scaling, delivering gains up to +8.2\%. ACE's higher total consumption (4.5--4.74M vs.\ 2.5--3.8M for ReasoningBank and ReMe) stems from its two-stage reflection-curation process and injecting all retrieved memories into the ReAct loop, compounding both new and cached token costs.

RefCon's efficiency comes from where its costs are concentrated. Trajectory generation, the dominant phase by volume, is cache-friendly due to structural redundancy across ReAct turns, keeping new token costs low throughout the generation phase. The additional overhead over sequential scaling is localized almost entirely to the extraction step, which processes three full trajectories of fresh, non-overlapping content and yields predominantly cache misses. This separation makes RefCon's trade-off favorable: trajectory generation via self-refinement is no more costly than sequential scaling, while the one-time extraction overhead purchases a meaningful jump in memory quality by consolidating all three refined trajectories, spending compute precisely where richer input improves downstream accuracy.

\subsection{Scaling Law}

\begin{figure}[t]
  \centering
  \includegraphics[width=\columnwidth]{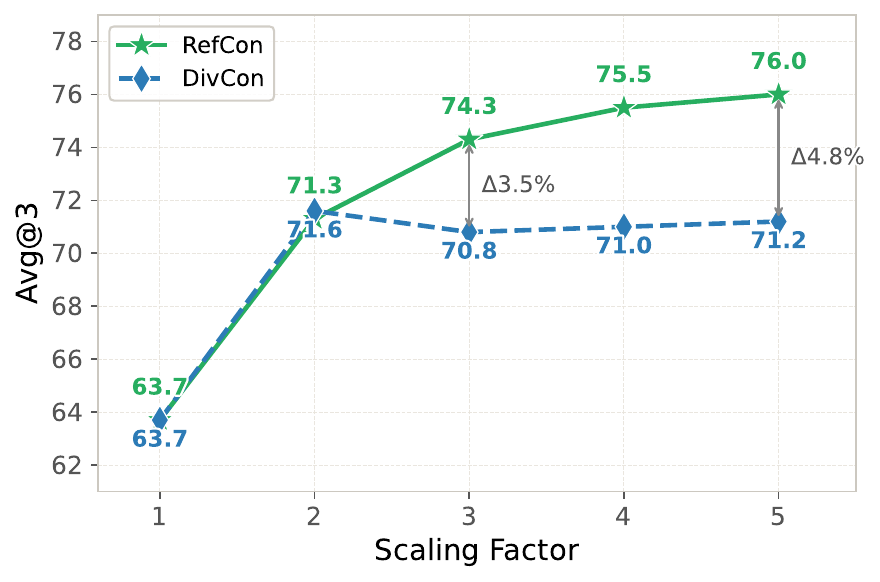}
  \caption{Scaling of RefCon and DivCon.}
  \label{fig:scaling-refcon}
\end{figure}

As shown in Figure~\ref{fig:scaling-refcon}, RefCon and DivCon share a baseline at $k=1$ (63.7\%) and achieve nearly identical performance at $k=2$ ($\approx$71.5\%), but diverge sharply thereafter: RefCon exhibits steady gains reaching 76.0\% at $k=5$, while DivCon plateaus around 71.0\%, suggesting additional diversity-seeking trajectories introduce conflicting noise beyond the initial gain. This indicates that while an initial round of diversity exploration provides a useful inductive signal, consistent refinement is more effective at converting increased compute into sustained accuracy. Solution quality proves a more reliable driver than solution diversity as trajectories scale. RefCon is therefore the more scalable strategy under larger compute budgets, as each additional trajectory reliably improves memory consolidation input rather than introducing variance the extractor must filter out, making it particularly well-suited to settings where inference compute can be scaled up.


\section{Conclusion}
We introduced RefCon, a memory-aware test-time scaling framework that unifies sequential self-refinement with parallel self-contrast to extract higher-quality memories for context-evolving agents without relying on gold labels. Notably, RefCon's label-free extraction surpasses some ground-truth baselines and maintains its effectiveness across different model scales. Across AppWorld and BFCL-V3, RefCon consistently delivers the strongest or near-strongest performance and favorable accuracy-token trade-offs, while our DivCon variant highlights when explicit diversity can help but also reveals stability and scaling limits. The successful application of RefCon to the SWE-bench benchmark further validates its generalizability to intricate, real-world software engineering environments. By capturing recurring routines and comparing trajectories, our approach enables effective memory extraction even in long-horizon tasks where distinguishing correct from incorrect reasoning is inherently difficult. These findings emphasize that improving memory quality at test time is a practical, compute-efficient path to continual agent improvement, and they motivate further work on more robust diversity generation and adaptive scaling strategies.

\section*{Limitations}
Our study has two main limitations. First, to address potential biases in model scale, we conducted experiments using both GLM-4.6 (large), Gemma 4 31B (medium), Qwen3.5 9B (small). Although incorporating three scales reduces the limitation of a single-model study, the generalizability of our results to models with different architectures or tuning styles is not yet fully guaranteed. We aim to broaden the scope of our evaluations to include more diverse model families in future iterations. Second, our test-time scaling analysis only evaluates up to five trajectories ($k\leq5$), which means we cannot yet characterize behavior under larger compute budgets, including whether performance continues to improve, plateaus, or degrades beyond this range.

\section*{Ethical Considerations}
This work does not raise specific ethical concerns. Our contributions focus on developing memory extraction frameworks for effective context-evolving agent. All experiments are conducted on publicly available benchmarks with open-source models, without involving human subjects, sensitive data, or privacy-related information. No potential conflicts of interest are present. While there is a potential for generated memories to occasionally degrade model performance, this occurrence is infrequent and can be mitigated through proper extraction or retrieval thresholds.


\bibliography{custom}

\appendix

\section*{Appendix}
\label{sec:appendix}

\section{RefCon and DivCon Detailed Results}
\label{sec:appendix-refcon-divcon}
Tables \ref{tab:gemma-results} and \ref{tab:qwen-results} present the performance of context-evolving agents using smaller language model architectures, specifically Gemma 4 31B and Qwen3.5 9B. By replicating our main results from Table \ref{tab:main-results} (GLM 4.6) at these scales, we provide a focused comparison across the most competitive methods. We observed a significant performance decline on the AppWorld benchmark when using Qwen3.5 9B, whereas the performance drop on BFCL-V3 remained marginal. In settings without ground-truth labels, RefCon consistently outperformed other context-evolving agents across both model scales, even surpassing ACE (with ground-truth) on AppWorld. However, these gains were less pronounced than those achieved with the larger GLM-4.6 model, suggesting that while RefCon’s scaling remains effective, smaller models still struggle with the nuances of high-fidelity memory extraction and reuse. Notably, some context-evolving methods even degraded performance on AppWorld when applied to Qwen3.5 9B. Gemma 4 31B has significantly better performance, similar to GLM 4.6. The benefit of memory reuse is also more aparrent in this model compared to Qwen3.5 9B, showing the importance of having strong base model. The underperformance of DivCon relative to RefCon further suggests that exploring diversity through prompting alone remains a significant challenge for smaller models, reinforcing the necessity of RefCon's structured refinement and contrastive approach.

In Table~\ref{tab:main-results} and Table~\ref{tab:matts-appworld}, we report Avg@3 for both RefCon and DivCon. For RefCon, the reported Avg@3 comes from the third run (after the second refinement), whereas for DivCon it comes from the first run. We execute RefCon three times, yielding first-, second-, and final-run results for each trial; the same setup is used for DivCon across its first and second runs. Because RefCon applies self-refinement, later runs can improve accuracy (Table~\ref{tab:refcon-detail}). By contrast, DivCon emphasizes diversity, which does not always lead to better scores (Table~\ref{tab:divcon-detail}). Therefore, we report the final run for RefCon and the first run for DivCon for all experiments.

\begin{figure}[t]
  \centering
  \includegraphics[width=\columnwidth]{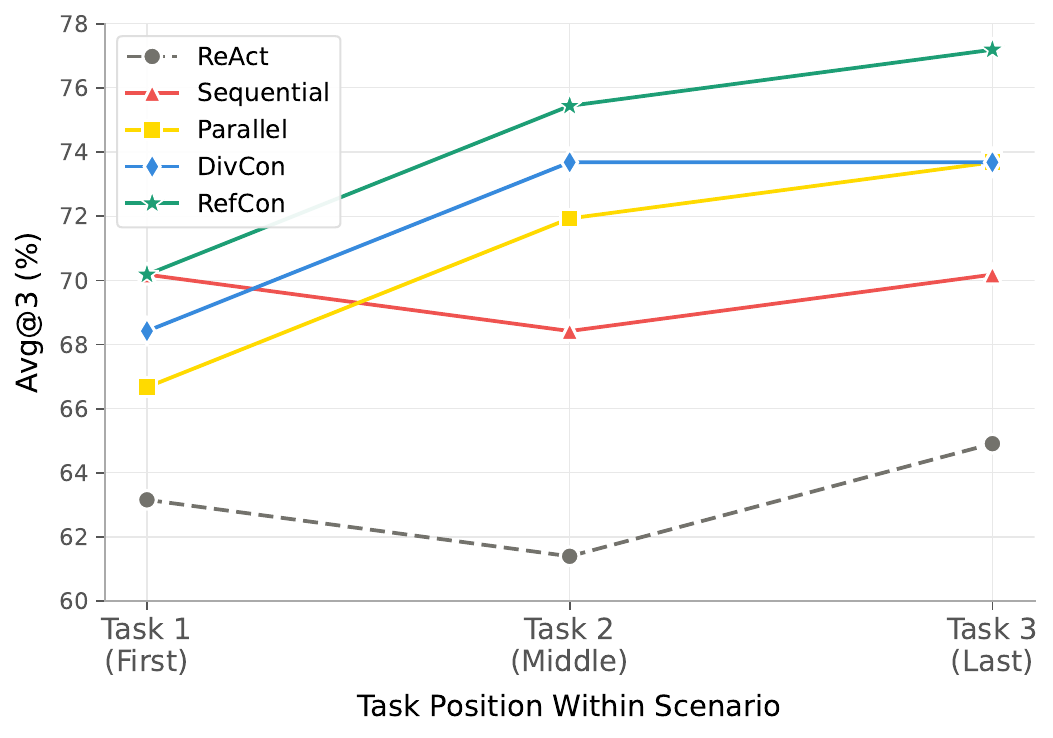}
  \caption{Avg@3 improvement across similar task within the same scenario on AppWorld sorted by chronological run order.}
  \label{fig:task-improvement}
\end{figure}

We further analyze the results in Table~\ref{tab:matts-appworld} (ReMe) in Figure~\ref{fig:task-improvement} to demonstrate RefCon’s superior ability to resolve \textbf{recurring} problems. The AppWorld dataset consists of three related tasks within a single scenario, varying only in their environmental states. By grouping these tasks according to their chronological execution order, we observe that all MaTTS methods demonstrate performance gains in subsequent tasks, indicating successful memory transfer. Notably, RefCon, DivCon, and Parallel scaling exhibit the steepest improvement curves between the first and second tasks, suggesting that the high-quality memory generated via self-contrast is instrumental in solving related tasks. While RefCon and Sequential scaling both achieve high initial performance on the first task due to self-refinement, Sequential scaling fails to maintain consistent improvements in later tasks, likely due to the limitations of extracting memory from a single trajectory.

\begin{table*}[!t]
\centering
\footnotesize
\caption{Performance of context-evolving agent methods on AppWorld and BFCL-V3 Using Gemma 4 31B.}
\label{tab:gemma-results}
\begin{tabular}{l cccc cc}
\toprule
\multirow{3}{*}{\textbf{Methods}} & \multicolumn{4}{c}{\textbf{AppWorld}} & \multicolumn{2}{c}{\textbf{BFCL-V3}} \\
\cmidrule(lr){2-5} \cmidrule(lr){6-7}
& \multicolumn{2}{c}{Avg@3} & \multicolumn{2}{c}{Pass@3} & Avg@3 & Pass@3 \\
\cmidrule(lr){2-3} \cmidrule(lr){4-5}
& TGC & SGC & TGC & SGC &  &  \\
\midrule
\rowcolor{gray!15}\multicolumn{7}{c}{Baseline} \\
ReAct & $73.68^{\pm 2.05}$ & $45.61^{\pm 2.48}$ & 92.98 & 78.95 & $61.33^{\pm 0.94}$ & 72.00 \\
\midrule
\rowcolor{gray!15}\multicolumn{7}{c}{With Gold Labels/Ground-Truth} \\
ACE (with Ground-Truth)      & $\textbf{80.12}^{\pm \textbf{2.19}}$ & $\textbf{54.39}^{\pm \textbf{4.96}}$ & \textbf{96.49} & \textbf{89.47} & $\textbf{69.33}^{\pm \textbf{3.40}}$ & \textbf{84.00} \\
ReMe (Sequential Scaling)    & $78.36^{\pm 3.51}$ & $51.46^{\pm 3.72}$ & 94.74 & 84.21 & $66.67^{\pm 2.49}$ & 80.00 \\
ReMe (Parallel Scaling)      & $76.02^{\pm 2.92}$ & $48.54^{\pm 4.96}$ & 92.98 & 84.21 & $64.67^{\pm 4.11}$ & 78.00 \\
\midrule
\rowcolor{gray!15}\multicolumn{7}{c}{Without Gold Labels/Ground-Truth} \\
ACE (without Ground-Truth)              & $74.27^{\pm 2.48}$ & $46.20^{\pm 2.48}$ & 91.23 & 73.68 & $62.00^{\pm 2.05}$ & 74.00 \\
ReasoningBank                           & $73.10^{\pm 3.51}$ & $44.74^{\pm 3.72}$ & 89.47 & 73.68 & $60.67^{\pm 2.83}$ & 72.00 \\
ReasoningBank (Sequential Scaling)      & $76.02^{\pm 2.63}$ & $48.54^{\pm 3.51}$ & 92.98 & 78.95 & $64.00^{\pm 2.49}$ & 76.00 \\
ReasoningBank (Parallel Scaling)        & $75.20^{\pm 3.07}$ & $47.37^{\pm 4.38}$ & 91.23 & 78.95 & $63.33^{\pm 3.30}$ & 74.00 \\
DivCon  & $76.61^{\pm 3.94}$ & $50.29^{\pm 3.72}$ & 94.74 & 84.21 & $65.33^{\pm 3.30}$ & 80.00 \\
RefCon  & $\textbf{79.53}^{\pm \textbf{4.38}}$ & $\textbf{53.22}^{\pm \textbf{5.42}}$ & \textbf{96.49} & \textbf{89.47} & $\textbf{68.00}^{\pm \textbf{5.25}}$ & \textbf{84.00} \\
\bottomrule
\end{tabular}
\end{table*}

\begin{table*}[!t]
\centering
\footnotesize
\caption{Performance of context-evolving agent methods on AppWorld and BFCL-V3 Using Qwen3.5 9B.}
\label{tab:qwen-results}
\begin{tabular}{l cccc cc}
\toprule
\multirow{3}{*}{\textbf{Methods}} & \multicolumn{4}{c}{\textbf{AppWorld}} & \multicolumn{2}{c}{\textbf{BFCL-V3}} \\
\cmidrule(lr){2-5} \cmidrule(lr){6-7}
& \multicolumn{2}{c}{Avg@3} & \multicolumn{2}{c}{Pass@3} & Avg@3 & Pass@3 \\
\cmidrule(lr){2-3} \cmidrule(lr){4-5}
& TGC & SGC & TGC & SGC &  &  \\
\midrule
\rowcolor{gray!15}\multicolumn{7}{c}{Baseline} \\
ReAct & $18.71^{\pm 2.19}$ & $10.53^{\pm 4.30}$ & 29.82 & 15.79 & $53.33^{\pm 3.40}$ & 72.00 \\
\midrule
\rowcolor{gray!15}\multicolumn{7}{c}{With Gold Labels/Ground-Truth} \\
ACE (with Ground-Truth) & $\textbf{18.13}^{\pm \textbf{4.60}}$ & $\textbf{8.77}^{\pm \textbf{2.48}}$ & \textbf{29.82} & \textbf{15.79} & $\textbf{56.00}^{\pm \textbf{4.32}}$ & \textbf{80.00} \\
ReMe (Sequential Scaling) &  $17.54^{\pm 2.48}$ & $7.89^{\pm 2.48}$  & 26.32 & \textbf{15.79} & $54.00^{\pm 3.27}$ & 76.00 \\
ReMe (Parallel Scaling) & $16.96^{\pm 3.51}$ & $7.02^{\pm 2.48}$  & 24.56 & 10.53 & $52.00^{\pm 3.27}$ & 74.00 \\
\midrule
\rowcolor{gray!15}\multicolumn{7}{c}{Without Gold Labels/Ground-Truth} \\
ACE (without Ground-Truth) &  $15.20^{\pm 2.48}$ & $5.85^{\pm 2.48}$  & 21.05 & 10.53 & $40.67^{\pm 3.40}$ & 60.00 \\
ReasoningBank & $14.04^{\pm 2.48}$ & $4.68^{\pm 2.48}$  & 19.30 & 10.53 & $37.33^{\pm 2.49}$ & 54.00 \\
ReasoningBank (Sequential Scaling) & $16.96^{\pm 2.48}$ & $8.77^{\pm 4.96}$ & 22.81 & 15.79 & $45.33^{\pm 0.94}$  &  68.00 \\
ReasoningBank (Parallel Scaling) & $17.54^{\pm 2.48}$ & $12.28^{\pm 4.96}$ & 22.81 & 15.79 & $43.33^{\pm 5.25}$ & 62.00 \\
DivCon & $16.96^{\pm 4.38}$ & $12.28^{\pm 2.48}$ & 26.32 & \textbf{21.05} & $46.00^{\pm 4.90}$ & 72.00 \\
RefCon & $\textbf{19.88}^{\pm \textbf{5.42}}$ & $\textbf{12.28}^{\pm \textbf{4.96}}$ & \textbf{31.58} & \textbf{21.05} & $\textbf{50.67}^{\pm \textbf{8.38}}$ & \textbf{80.00} \\
\bottomrule
\end{tabular}
\end{table*}

\begin{table*}[t]
\centering
\small
\caption{RefCon accuracy across runs for each method.}
\label{tab:refcon-detail}
\begin{tabular}{l cccc}
\toprule
\multirow{3}{*}{\textbf{Methods}} & \multicolumn{4}{c}{\textbf{AppWorld}} \\
\cmidrule(lr){2-5}
& \multicolumn{2}{c}{Avg@3} & \multicolumn{2}{c}{Pass@3} \\
\cmidrule(lr){2-3} \cmidrule(lr){4-5} 
& TGC & SGC & TGC & SGC \\
\midrule
\rowcolor{gray!15}\multicolumn{5}{c}{ACE} \\
First Run & $68.40^{\pm 5.72}$ & $\textbf{43.87}^{\pm \textbf{6.57}}$ & 85.96 & 68.42 \\
Second Run (First Refinement) & $70.17^{\pm 6.57}$ & $42.1^{\pm 8.57}$ & \textbf{92.98} & \textbf{84.21} \\
Third Run (Second Refinement) & $\textbf{72.53}^{\pm \textbf{5.82}}$ & $\textbf{43.87}^{\pm \textbf{6.57}}$ & \textbf{92.98} & 78.95 \\
\midrule
\rowcolor{gray!15}\multicolumn{5}{c}{ReasoningBank} \\
First Run & $70.76^{\pm 0.80}$ & $47.37^{\pm 8.61}$ & 89.47 & \textbf{84.21} \\
Second Run (First Refinement) & $\textbf{71.93}^{\pm \textbf{2.86}}$ & $\textbf{50.87}^{\pm \textbf{10.85}}$ & 91.23 & \textbf{84.21} \\
Third Run (Second Refinement) & $\textbf{71.93}^{\pm \textbf{3.80}}$ & $49.13^{\pm 13.13}$ & \textbf{92.98} & \textbf{84.21} \\
\midrule
\rowcolor{gray!15}\multicolumn{5}{c}{ReMe} \\
First Run & $69.59^{\pm 2.98}$ & $50.87^{\pm 2.48}$ & 89.47 & \textbf{78.95} \\
Second Run (First Refinement) & $71.35^{\pm 4.38}$ & $49.13^{\pm 2.48}$ & 91.23 & \textbf{78.95} \\
Third Run (Second Refinement) & $\textbf{74.30}^{\pm \textbf{0.94}}$ & $\textbf{52.67}^{\pm \textbf{7.45}}$ & \textbf{92.98} & \textbf{78.95} \\
\bottomrule
\end{tabular}
\end{table*}

\begin{table*}[t]
\centering
\small
\caption{DivCon accuracy across runs for each method.}
\label{tab:divcon-detail}
\begin{tabular}{l cccc}
\toprule
\multirow{3}{*}{\textbf{Methods}} & \multicolumn{4}{c}{\textbf{AppWorld}} \\
\cmidrule(lr){2-5}
& \multicolumn{2}{c}{Avg@3} & \multicolumn{2}{c}{Pass@3} \\
\cmidrule(lr){2-3} \cmidrule(lr){4-5} 
& TGC & SGC & TGC & SGC \\
\midrule
\rowcolor{gray!15}\multicolumn{5}{c}{ACE} \\
First Run & $\textbf{69.00}^{\pm \textbf{2.16}}$ & $\textbf{42.10}^{\pm \textbf{4.33}}$ & \textbf{91.23} & \textbf{78.95}  \\
Second Run (First Alternative) & $65.50^{\pm 2.98}$ & $40.37^{\pm 6.57}$ & 89.47 & 73.68 \\
\midrule
\rowcolor{gray!15}\multicolumn{5}{c}{ReasoningBank} \\
First Run & $\textbf{76.03}^{\pm \textbf{2.21}}$ & $\textbf{52.67}^{\pm \textbf{0.00}}$ & \textbf{94.74} & \textbf{89.47} \\
Second Run (First Alternative) & $70.17^{\pm 2.50}$ & $47.33^{\pm 7.45}$ & 92.98 & \textbf{89.47} \\
\midrule
\rowcolor{gray!15}\multicolumn{5}{c}{ReMe} \\
First Run & $\textbf{71.93}^{\pm \textbf{5.76}}$ & $\textbf{47.37}^{\pm \textbf{4.29}}$ & 92.98 & 78.95 \\
Second Run (First Alternative) & $67.25^{\pm 2.19}$ & $38.60^{\pm 2.48}$ & \textbf{96.49} & \textbf{89.47} \\
\bottomrule
\end{tabular}
\end{table*}

\section{Memory Examples}
\label{sec:appendix-example}

We use ReAct framework as base and utilize ACE, ReasoningBank, and ReMe as baseline context-evolving agents.  ACE \citep{zhang2025ACE} extracts and curates memories through a reflector–curator agent pipeline but directly uses all stored memories without a dedicated retrieval mechanism. ReasoningBank \citep{ouyang2025ReasoningBank} introduces the MaTTS paradigm (both parallel and sequential), which performs memory extraction from both successful and failed trajectories and retrieves memories using embedding-based similarity. ReMe \citep{cao2025ReMe} further improves retrieval by incorporating a retrieve–rerank–rewrite pipeline to filter and refine retrieved memories. These methods also differ in their test-time scaling strategies, including parallel and sequential trajectory refinement. ACE doesn't have default test-time scaling strategies, while ReMe has different parallel and sequential scaling strategy compared to ReasoningBank. For parallel, it extracts memories from best, worst, and comparison of best-worst trajectories. While for sequential scaling, it only retries failed trajectory and use it as cautionary memory for the next generation.

ACE, ReasoningBank, and ReMe use different memory schemas. ACE stores each memory as a \textit{section} and \textit{content} pair (Table~\ref{tab:ace-memory-example}); the section acts as a stable bucket (e.g., strategy, tool usage, or caution) so related memories can be organized and retrieved consistently. ReasoningBank represents memory with \textit{title}, \textit{description}, and \textit{content} (Table~\ref{tab:reasoningbank-memory-example}); this structure separates a short identifier (title), a concise summary for quick screening (description), and the full actionable detail (content), which is useful when comparing many candidate memories during retrieval. ReMe uses \textit{when\_to\_use} and \textit{content} (Table~\ref{tab:reme-memory-example}); this emphasizes direct applicability by explicitly encoding the trigger condition first, followed by the action or rule to execute. Furthermore, ReMe retrieved memories will be rewritten into a coherent step-by-step guide paragraph. All examples are extracted from trajectories generated for the same query: "How many unique songs are there across my Spotify song library, albums library and all playlists?" (\textit{fac291d\_1}) The underlying agent frameworks differ, but all examples use the same scaling method, RefCon.

\section{Example of Successful RefCon Improvement}
\label{sec:appendix-improvement}
This section presents a task that RefCon solves but sequential scaling fails on. For analysis, we compare retrieved memories (Table~\ref{tab:retrieved-memory-comparison}) and trajectory snippets (Table~\ref{tab:trajectory-diff}). The task is \textit{0d8a4ee\_2} with the query: "Send the following phone message to my siblings and roommates, who do not have a venmo account, "Get on venmo please!". Although each method retrieves five memories, we display only two that differ and materially influence trajectory generation. The memories retrieved by sequential scaling push the agent to skip verification and jump directly to sending messages. In contrast, RefCon retrieves memories that encourage API exploration and case checking. This suggests that RefCon extracts more useful memories by self-contrasting across multiple trajectories, while sequential scaling extracts weak guidance from a single (possibly failed) trajectory. The effect is visible in the rollouts: sequential scaling brute-forces API field combinations, fails, and then sends messages to \texttt{all\_contact}. On the other hand, RefCon first inspects the data structure, then refines the code and correctly identifies \texttt{contacts\_without\_venmo}.

Table~\ref{tab:trajectory-diverging} illustrates how a single API misuse in Step~12 propagates into a completely wrong final answer. Both trajectories follow identical steps up to and including Step~11, correctly collecting 79 unique song IDs and filtering to 18 R\&B candidates. The divergence arises when retrieving play counts: the successful trajectory reads \texttt{play\_count} directly from \texttt{show\_song}, which exposes it as part of the public song schema, while the failed trajectory makes an additional call to \texttt{show\_song\_privates} under the plausible assumption that per-user listening data would reside there. However, \texttt{show\_song\_privates} only returns interaction flags (\texttt{liked}, \texttt{reviewed}, \texttt{in\_song\_library}, \texttt{downloaded}), and contains no \texttt{play\_count} field. The \texttt{.get('play\_count',~0)} fallback silently assigns a count of zero to all 18 songs, so the subsequent sort in Step~13 is applied to a uniform list and produces an arbitrary ordering. Crucially, the failed trajectory raises no error at any point, the code is syntactically valid, the API calls succeed, and the output is well-formed.
It makes class of mistake particularly difficult to detect without careful cross-referencing of API schemas. This case is where self-contrast reasoning in RefCon shows its benefit.

\section{Hyperparameter Details}
\label{sec:appendix-hyperparams}
We generally employ a scaling factor of 3 (except 2 for DivCon). To promote trajectory diversity, we use parallel rollouts with varying temperatures \citep{holtzman2020temp,wang2023temp}. Across all methods, we prompt the LLM to extract at most five memories per extraction step. For parallel scaling, we rollout three trajectories at temperatures of 0.7, 0.85, and 1.0 to ensure diversity; otherwise, the default temperature is 0.7. For RefCon and ReMe, we utilize a top-$k$ retrieval and reranking strategy with specific usage and utility pruning thresholds. Method-specific configuration are as follows:
\begin{itemize}
\item \textbf{ReasoningBank}: Uses top-1 query retrieval, returning up to five memories for each query.
\item \textbf{ACE}: The playbook is capped at 50 memories, and ground-truth usage is disabled during scaling.
\item \textbf{ReMe \& RefCon}: We use a top-$k$ retrieval of 10, the only choose top-5 after reranking. We set a deduplication similarity threshold $\epsilon = 0.5$, and pruning thresholds $\alpha = 5n$ (usage) and $\beta = 0.5$ (utility), where $n$ represents the scaling factor.
\end{itemize}

\section{Prompt Details}
\label{sec:appendix-prompt}
Prompt details for memory extraction and memory retrieval for each method: ACE, ReasoningBank, and ReMe can be found on the original paper. We only modify the number of trajectory provided in the prompt template for RefCon, DivCon, and parallel scaling, while the instruction remains the same. For memory extraction, RefCon adopts ReasoningBank's self-contrast reasoning and ReMe's memory format as presented in Table~\ref{tab:refcon-prompt}. We also provide the self-refine and self-diversity prompts for AppWorld in Table~\ref{tab:appworld-prompt}, and the corresponding prompts for BFCL-V3 in Table~\ref{tab:bfcl-prompt}. In our experiments, applying the simpler AppWorld-style refinement prompt to BFCL-V3 was ineffective, since the trajectories showed little to no change. To address this, we include both a critique prompt and an ideation prompt to better guide trajectory refinement, as shown in Table~\ref{tab:bfcl-prompt}. However in some cases, self-refinement can even degrade trajectories that were already correct. We also include prompt for SWE-bench Verified (Mini) refinement in Table~\ref{tab:swe-prompt}.

\begin{table}[t]
\centering
\small
\caption{Example of memories extracted using ACE with RefCon scaling.}
\label{tab:ace-memory-example}
\begin{tabular}{p{0.95\linewidth}}
\toprule
\textbf{Section:} \texttt{strategies\_and\_hard\_rules} \\
\textbf{Content:} For Spotify authentication, always call \texttt{apis.supervisor.show\_account\_passwords()} to retrieve stored credentials instead of creating dummy passwords. \\
\\[0.2em]
\textbf{Section:} \texttt{strategies\_and\_hard\_rules} \\
\textbf{Content:} When collecting songs across multiple libraries (songs, albums, playlists), use \texttt{set()} to deduplicate \texttt{song\_ids} from all sources before processing. \\
\\[0.2em]
\textbf{Section:} \texttt{strategies\_and\_hard\_rules} \\
\textbf{Content:} Music library query workflow: 1) Authenticate \textrightarrow{} 2) Collect song IDs from all libraries (songs/albums/playlists) \textrightarrow{} 3) Deduplicate using \texttt{set()} \textrightarrow{} 4) Get song details via \texttt{show\_song} for each ID \textrightarrow{} 5) Filter by criteria (genre, etc.) \textrightarrow{} 6) Sort by \texttt{play\_count} \textrightarrow{} 7) Format output. \\
\\[0.2em]
\textbf{Section:} \texttt{apis\_to\_use\_for\_specific\_information} \\
\textbf{Content:} For Spotify authentication, always call \texttt{apis.supervisor.show\_account\_passwords()} to retrieve stored credentials instead of creating dummy passwords. \\
\bottomrule
\end{tabular}
\end{table}

\begin{table}[!t]
\centering
\small
\caption{Example of memories extracted using ReasoningBank with RefCon scaling.}
\label{tab:reasoningbank-memory-example}
\begin{tabular}{p{0.95\linewidth}}
\toprule
\textbf{Title:} Authenticate Before Accessing Private Data \\
\textbf{Description:} Always authenticate with the service before attempting to access private user data like libraries. \\
\textbf{Content:} When working with music streaming services, private library data (songs, albums, playlists) requires authentication. First explore available APIs, then use stored credentials to obtain an access token before making any data retrieval requests. \\
\\[0.2em]
\textbf{Title:} Handle Pagination for Complete Data Retrieval \\
\textbf{Description:} Implement pagination loops to retrieve all items from paginated API responses. \\
\textbf{Content:} Many APIs return data in pages with limits. Use while loops that increment \texttt{page\_index} until an empty response is returned, collecting all items across pages to ensure complete data retrieval. \\
\\[0.2em]
\textbf{Title:} Extract Song IDs from Different Library Types \\
\textbf{Description:} Different library types require different approaches to extract song identifiers. \\
\textbf{Content:} Song libraries directly return song objects with \texttt{song\_id} fields, while album libraries return albums containing \texttt{song\_ids} arrays, and playlist libraries return playlists with \texttt{song\_ids} arrays. Extract and flatten these appropriately. \\
\\[0.2em]
\textbf{Title:} Use Sets for Unique Counting \\
\textbf{Description:} Use set data structures to automatically handle uniqueness when counting items. \\
\textbf{Content:} After collecting all song IDs from various sources, add them to a set which automatically handles duplicates. The final count is simply the length of this set. \\
\bottomrule
\end{tabular}
\end{table}

\begin{table*}[t]
\centering
\small
\caption{Example of memories extracted using ReMe with RefCon scaling.}
\label{tab:reme-memory-example}
\begin{tabular}{p{0.95\linewidth}}
\toprule
\textbf{Original Memory} \\
\midrule
\textbf{When To Use:} When working with paginated API responses\\
\textbf{Content:} All trajectories successfully implemented pagination with while loops and \texttt{page\_index} incrementing. This consistent pattern across all successful trajectories shows that proper pagination handling is critical for complete data retrieval. The pattern of checking for empty responses to break the loop is reliable. \\
\\[0.2em]
\textbf{When To Use:} When handling authentication tokens in multi-step workflows \\
\textbf{Content:} All trajectories successfully extracted and reused the \texttt{access\_token} from login responses across multiple API calls. The consistent pattern of storing \texttt{spotify\_access\_token = login\_result['access\_token']} and passing it to subsequent calls demonstrates proper token management. None of the trajectories encountered token expiration issues, suggesting the token lifespan was sufficient for the workflow duration. \\
\\[0.2em]
\textbf{When To Use:} When handling data deduplication across multiple sources \\
\textbf{Content:} All trajectories correctly used sets for deduplication, but Trajectory 3 demonstrated the most elegant approach by updating the set with comprehensions within the helper function. Trajectories 1, 2, and 4 used more verbose loop structures. The set-based approach was universally successful, proving it's the right pattern for this type of unique counting task across multiple data sources. \\
\\[0.2em]
\textbf{When To Use:} When deciding whether to explore API documentation before implementation \\
\textbf{Content:} Trajectory 3 successfully skipped explicit API documentation exploration and went directly to implementation, demonstrating that prior knowledge or confidence in API structure can eliminate unnecessary exploration steps. However, Trajectories 1, 2, and 4 took a more cautious approach by examining API docs first, which is safer for unfamiliar APIs but adds overhead. The doc exploration didn't change the implementation approach, suggesting it was redundant in this case. \\
\midrule
\textbf{Rewritten Memory} \\
\midrule
To accurately count unique songs across your Spotify library, you'll need to implement a comprehensive data aggregation approach that handles multiple library sources with potential overlaps. \textbf{Start by authenticating} with Spotify's API using stored credentials, \textbf{then systematically explore the available endpoints} for songs, albums, and playlists. \textbf{The key is to use a set data structure} to track unique song IDs as you paginate through each library source. \textbf{Implement while loops} to handle pagination for songs library, albums library, and all playlists separately, adding each song ID to your set which automatically handles deduplication across all sources. This ensures you capture every unique song exactly once, even if the same track appears in multiple playlists or albums. After processing all paginated results from all three library types, the final count is simply the size of your unique song IDs set. This approach efficiently handles the overlapping nature of Spotify's library structure while providing an accurate count of your total unique songs. \\
\bottomrule
\end{tabular}
\end{table*}

\begin{table*}[t]
\centering
\small
\caption{Retrieved memory comparison of sequential (failed) and RefCon (success) scaling.}
\label{tab:retrieved-memory-comparison}
\begin{tabular}{p{0.95\linewidth}}
\toprule
\textbf{Retrieved Memory (Sequential Scaling)} \\
\midrule
\textbf{When To Use:} When you encounter verification issue in Venmo \\
\textbf{Content:} If you encounter issues with optional verification steps (like checking Venmo status), proceed with the main task (sending messages) to ensure completion. The primary goal of delivering the message is more important than perfect filtering, especially when verification may be unreliable. \\
\\[0.2em]
\textbf{When To Use:} When filtering contacts by a condition and finding no matches, but the user has requested to send a specific message \\
\textbf{Content:} If no recipients match the specified filter criteria, consider alternative interpretations of the user's request before concluding the task. The message content itself may provide clues about the intended audience. \\
\midrule
\textbf{Retrieved Memory (RefCon)} \\
\midrule
\textbf{When To Use:} When filtering contact on Venmo \\
\textbf{Content:} Across trajectories that attempted Venmo filtering, the reliable pattern was to query users by email first, then validate exact email matches (case-insensitive) before excluding a contact from messaging. Trajectories that treated any non-empty search response as a match risked false positives, while trajectories that performed exact email verification were more robust. In this setup, an empty search result consistently signaled no Venmo account. \\
\\[0.2em]
\textbf{When To Use:} When using Venmo API for sending money \\
\textbf{Content:} Trajectory 1 initially failed trying to use \texttt{receiver\_phone} parameter, then succeeded with \texttt{receiver\_email}. All successful trajectories used \texttt{receiver\_email} from contact information. PRINCIPLE: Always check API documentation for required parameters---Venmo specifically requires \texttt{receiver\_email}, not \texttt{receiver\_phone}, even when you have the phone number. This also applies to any endpoint when you failed to use it properly. \\
\bottomrule
\end{tabular}
\end{table*}

\begin{table*}[t]
\centering
\small
\caption{Trajectory comparison with memory retrieved from sequential (failed) and RefCon (success) scaling.}
\label{tab:trajectory-diff}

\renewcommand{\arraystretch}{1.2}
\begin{tabularx}{\textwidth}{
  >{\cellcolor{failred}}X
  >{\cellcolor{passgreen}}X
}

\multicolumn{1}{c}{\colorbox{failheader}{\parbox{0.44\textwidth}{\centering\color{white}\bfseries\normalsize Sequential Scaling (Failed)}}}
&
\multicolumn{1}{c}{\colorbox{passheader}{\parbox{0.44\textwidth}{\centering\color{white}\bfseries\normalsize RefCon (Successful)}}} \\[4pt]

\toprule

$\ldots$\newline
\textbf{Assistant:}\newline
\texttt{contacts\_without\_venmo = []}\newline
\texttt{for contact in all\_contacts:}\newline
\texttt{\ \ email = contact['email']}\newline
\texttt{\ \ search\_result = apis.venmo.search\_users(}\newline
\texttt{\ \ \ \ access\_token=venmo\_access\_token,}\newline
\texttt{\ \ \ \ query=email, page\_limit=1)}\newline
\texttt{\ \ if not search\_result:}\newline
\texttt{\ \ \ \ contacts\_without\_venmo.append(contact)}\newline
\texttt{print(contacts\_without\_venmo)}
\Env{[]}
\textbf{Assistant:}\newline
\texttt{contacts\_without\_venmo = []}\newline
\texttt{for contact in all\_contacts:}\newline
\texttt{\ \ email = contact['email']}\newline
\texttt{\ \ search\_result = apis.venmo.search\_users(}\newline
\texttt{\ \ \ \ access\_token=venmo\_access\_token,}\newline
\texttt{\ \ \ \ query=email)}\newline
\texttt{\ \ if not search\_result:}\newline
\texttt{\ \ \ \ contacts\_without\_venmo.append(contact)}\newline
\texttt{print(contacts\_without\_venmo)}
\Env{[]}
\textbf{Assistant:}\newline
\texttt{contacts\_without\_venmo = []}\newline
\texttt{for contact in all\_contacts:}\newline
\texttt{\ \ full\_name = f"\{contact['first\_name']\} \{contact['last\_name']\}"}\newline
\texttt{\ \ search\_result = apis.venmo.search\_users(}\newline
\texttt{\ \ \ \ access\_token=venmo\_access\_token,}\newline
\texttt{\ \ \ \ query=full\_name)}\newline
\texttt{\ \ if not search\_result:}\newline
\texttt{\ \ \ \ contacts\_without\_venmo.append(contact)}\newline
\texttt{print(contacts\_without\_venmo)}
\Env{[]}
$\ldots$

&

$\ldots$\newline
\textbf{Assistant:}\newline
\texttt{all\_contacts = siblings + roommates}\newline
\texttt{contacts\_without\_venmo = []}\newline
\texttt{for contact in all\_contacts:}\newline
\texttt{\ \ search\_result = apis.venmo.search\_users(}\newline
\texttt{\ \ \ \ access\_token=venmo\_access\_token,}\newline
\texttt{\ \ \ \ query=contact['email'])}\newline
\texttt{\ \ if not search\_result:}\newline
\texttt{\ \ \ \ contacts\_without\_venmo.append(contact)}\newline
\texttt{print(contacts\_without\_venmo)}
\Env{[]}
\textbf{Assistant:}\newline
\texttt{print(apis.venmo.search\_users(}\newline
\texttt{\ \ access\_token=venmo\_access\_token,}\newline
\texttt{\ \ query='kevin.blac@gmail.com'))}
\Env{[\{"first\_name": "Kevin", "email": "kevin.blac@gmail.com", \ldots\}]}
\textbf{Assistant:}\newline
\texttt{all\_contacts = siblings + roommates}\newline
\texttt{contacts\_without\_venmo = []}\newline
\texttt{for contact in all\_contacts:}\newline
\texttt{\ \ search\_result = apis.venmo.search\_users(}\newline
\texttt{\ \ \ \ access\_token=venmo\_access\_token,}\newline
\texttt{\ \ \ \ query=contact['email'])}\newline
\texttt{\ \ found = any(user['email'] == contact['email']}\newline
\texttt{\ \ \ \ for user in search\_result)}\newline
\texttt{\ \ if not found:}\newline
\texttt{\ \ \ \ contacts\_without\_venmo.append(contact)}\newline
\texttt{print(contacts\_without\_venmo)}
\Env{[\{"contact\_id": 801, "first\_name": "Martin", \ldots\}]}
$\ldots$

\\
\bottomrule
\end{tabularx}
\end{table*}

\begin{table*}[t]
\centering
\caption{Step-by-step comparison of the successful and failed trajectories.
Steps 1--11 are identical in both; the slight divergence occurs at Step~12,
propagating to a wrong final answer at Step~13.}
\label{tab:trajectory-diverging}
 
\renewcommand{\arraystretch}{1.35}
\begin{tabular}{
  >{\centering\arraybackslash}p{1cm}
  p{7cm}
  p{7cm}
}
\toprule
\textbf{Step} & \textbf{Successful Trajectory} & \textbf{Failed Trajectory} \\
\midrule
 
\rowcolor{lightgray}
1--2
& \multicolumn{2}{p{14.5cm}}{
  \textit{(Identical)} Retrieve account passwords; log in to Spotify to obtain \texttt{access\_token}.
} \\
 
\rowcolor{lightgray}
3--7
& \multicolumn{2}{p{14.5cm}}{
  \textit{(Identical)} Inspect API schemas for \texttt{show\_song}, \texttt{show\_song\_privates},
  \texttt{show\_song\_library}, \texttt{show\_album\_library}, \texttt{show\_playlist\_library}.
} \\
 
\rowcolor{lightgray}
8--11
& \multicolumn{2}{p{14.5cm}}{
  \textit{(Identical)} Collect 79 unique song IDs from all three libraries; filter to 18 R\&B songs via \texttt{show\_song()} genre check.
} \\
 
\midrule
 
\rowcolor{highlightbg}
\textbf{12}
&
\cellcolor{codepass}%
\textbf{Read \texttt{play\_count} from \texttt{show\_song}}

\smallskip
\begin{lstlisting}[style=plaintable]
rb_songs_with_play_count = []
for song_id in all_song_ids:
  song_info = apis.spotify
    .show_song(song_id=song_id)
  if song_info and 'r&b' in
      song_info['genre'].lower():
    rb_songs_with_play_count.append({
      'title': song_info['title'],
      'play_count': song_info['play_count']
    })
\end{lstlisting}

\smallskip
\texttt{show\_song} returns a rich public schema including \texttt{play\_count}:
{\footnotesize\texttt{\{song\_id, title, album\_id, duration,}}
\texttt{artists, genre, \underline{play\_count}, rating, ...\}}

\smallskip
Genre filtering and play count retrieval are done in a \textbf{single loop}. All 18 R\&B songs receive their correct play counts.

&
\cellcolor{codefail}%
\textbf{Read \texttt{play\_count} from \texttt{show\_song\_privates}}

\smallskip
\begin{lstlisting}[style=plaintable]
rb_songs_with_play_count = []
for song in rb_song_details:
  song_id = song['song_id']
  private_info = apis.spotify
    .show_song_privates(
      song_id=song_id,
      access_token=spotify_access_token)
  if private_info:
    play_count = private_info
      .get('play_count', 0)
    rb_songs_with_play_count.append({
      'title': song['title'],
      'play_count': play_count
    })
\end{lstlisting}

\smallskip
\texttt{show\_song\_privates} only exposes per-user interaction flags:
{\footnotesize\texttt{\{liked, reviewed,}}
\texttt{in\_song\_library, downloaded\}}

\smallskip
There is \textbf{no} \texttt{play\_count} field. The \texttt{.get('play\_count',\,0)} fallback silently returns \textbf{0 for all 18 songs}, producing no error and no warning.

\\
 
\midrule
 
\rowcolor{highlightbg}
\textbf{13}
&
\cellcolor{codepass}%
Sorted by real play counts:
 
\smallskip
\begin{lstlisting}[style=plaintable]
["Mysteries of the Silent Sea",
 "Crimson Veil",
 "Haunted Memories",
 "Fire and Ice"]
\end{lstlisting}
 
&
\cellcolor{codefail}%
All counts equal 0; order is arbitrary:
 
\smallskip
\begin{lstlisting}[style=plaintable]
["Shadows of the Past",
 "When Fate Becomes a Foe",
 "The Curse of Loving You",
 "Lost in a Moment's Grace"]
\end{lstlisting}
 
\\
 
\bottomrule
\end{tabular}
\end{table*}

\begin{table*}[t]
\centering
\small
\caption{Memory extraction prompt used for our context-evolving agent utilizing RefCon. We adopt ReMe memory format, but utilize self-contrast reasoning to extract memory.}
\label{tab:refcon-prompt}
\begin{tabular}{p{0.95\linewidth}}
\toprule
\textbf{Self-Contrast Prompt} \\
\midrule
You are an expert AI analyst comparing multiple step sequences which might be successful or failed to extract differential insights. \\
Your task is to compare and contrast these trajectories to identify the most useful and generalizable strategies as memory items using self-contrast reasoning. \\
Focus on critical decision points, technique variations, and approach differences. \\
COMPARATIVE ANALYSIS FRAMEWORK: \\
- DECISION CONTRAST: Compare critical decisions made in success vs failure cases \\
- TECHNIQUE VARIATIONS: Identify different approaches and their outcomes \\
- TIMING DIFFERENCES: Analyze when certain actions were taken and their impact \\
- SUCCESS FACTORS: Extract what specifically made the difference \\
EXTRACTION PRINCIPLES: \\
- Frame comparisons as PRINCIPLES as well as case-specific SOLUTIONS \\
- Identify PATTERNS that differentiate effective vs ineffective approaches \\
- Extract RULES that can guide future similar situations \\
- Focus on UNDERLYING MECHANISMS rather than surface-level differences \\
Trajectory 1:\\
\texttt{\{trajectory\_1\}} \\
\ldots \\
Trajectory N:\\
\texttt{\{trajectory\_n\}} \\
OUTPUT FORMAT: \\
Generate up to 5 comparative insights as JSON objects: \\
\texttt{[} \\
\texttt{\{} \\
\texttt{\ \ "when\_to\_use": "Specific scenarios where this comparative insight applies",} \\
\texttt{\ \ "experience": "Detailed comparison highlighting why success approach works better",} \\
\texttt{\ \ "tags": ["comparative\_analysis", "success\_factors", "relevant\_keywords"],} \\
\texttt{\ \ "confidence": 0.8,} \\
\texttt{\ \ "step\_type": "reasoning|action|observation|decision",} \\
\texttt{\ \ "tools\_used": ["list", "of", "tools"]} \\
\texttt{\}} \\
\texttt{]} \\
\bottomrule
\end{tabular}
\end{table*}

\begin{table*}[t]
\centering
\small
\caption{RefCon and DivCon trajectory generation prompt for AppWorld.}
\label{tab:appworld-prompt}
\begin{tabular}{p{0.95\linewidth}}
\toprule
\textbf{Self-Refine Prompt for AppWorld} \\
\midrule
Let's carefully re-examine the previous trajectory, including your reasoning steps and action taken. Pay special attention to whether you used the best API sequence and whether you used the API correctly. If you find inconsistencies, correct them. If everything seems correct, make it more efficient. Now, solve the same problem again from scratch. \\
\midrule
\textbf{Self-Diversity Prompt for AppWorld} \\
\midrule
Below is the previous trajectory, the solution might be correct or wrong. Now solve the same problem using a DIFFERENT reasoning approach. Focus on exploring alternative strategies. \\
\bottomrule
\end{tabular}
\end{table*}

\begin{table*}[t]
\centering
\small
\caption{RefCon and DivCon trajectory generation prompt for BFCL-V3.}
\label{tab:bfcl-prompt}
\begin{tabular}{p{0.95\linewidth}}
\toprule
\textbf{Critique Trajectory Prompt for Self-Refine} \\
\midrule
You are an expert reviewer analyzing an AI assistant's multi-turn tool-calling trajectory. Your job is to identify mistakes, missed actions, and suboptimal decisions. For each turn in the trajectory, evaluate: \\
1. Did the assistant call the appropriate tools? If a user requested an action (e.g., book, cancel, update), did the assistant actually make a tool call, or did it just respond with text? \\
2. Were the tool arguments correct? Check for wrong parameter values, missing required arguments, or arguments that contradict the user's request. \\
3. Did the assistant use information from previous tool responses correctly? For example, if a lookup returned an ID, did the assistant use that ID in subsequent calls? \\
4. Were there any unnecessary or redundant tool calls? \\
5. Did the assistant follow the logical sequence of operations? (e.g., lookup before booking, authenticate before accessing protected resources) \\
Be specific about which turns have issues and what should be done differently. If a turn looks correct, briefly note it as OK. Focus most of your analysis on turns that seem problematic. \\
IMPORTANT: You must respond ONLY with your critique. Do not attempt to solve the task yourself. \\
\midrule
\textbf{Self-Refine Prompt for BFCL-V3} \\
\midrule
You are an AI assistant tasked with completing a multi-turn tool-calling objective. A reviewer has analyzed your previous attempt and provided a critique. \\
1. Read the critique carefully to understand the mistakes made in the previous trajectory. \\
2. Trust the Schemas: The critique is a helpful guide, but your provided tool schemas are the absolute truth. If the critique suggests using a tool or parameter that does not exist in your schemas, ignore it. \\
3. Do not ask the user for any additional information/clarification, you are authorized to do everything in behalf of the user. \\
4. Restart the task from the beginning. Your environment state has been reset. \\
5. You are currently at the start of the task, but the provided previous trajectory covers the entire multi-turn interaction. Do NOT execute or simulate future steps ahead of time. Focus ONLY on the current task instruction. \\
6. You can use related memory to guide your reasoning to solve the problem. \\
\midrule
\textbf{Ideate Solution Prompt for Self-Diversity} \\
\midrule
You are an expert AI brainstorming assistant analyzing a previous interaction trajectory. Your job is to read the user's request and the previous approach, and then propose a DIFFERENT but valid reasoning approach or sequence of tool calls that solves the same problem. Focus your analysis on: \\
1. Understanding the user's core intent. \\
2. Identifying the strategy used in the previous trajectory. \\
3. Proposing alternative strategies, using different tools, or a different sequence of operations that could achieve the same goal successfully. \\
Be specific about the proposed alternative approach. \\
IMPORTANT: You must respond ONLY with your alternative idea. Do not attempt to solve the task yourself. \\
\midrule
\textbf{Self-Diversity Prompt for BFCL-V3} \\
\midrule
You are an AI assistant tasked with completing a multi-turn tool-calling objective. A reviewer has analyzed your previous attempt and provided an alternative idea for a different reasoning approach. \\
1. Read the alternative idea carefully and use it to solve the problem. \\
2. Trust the Schemas: The alternative idea is a helpful guide, but your provided tool schemas are the absolute truth. If the alternative idea suggests using a tool or parameter that does not exist in your schemas, ignore it. \\
3. Do not ask the user for any additional information/clarification, you are authorized to do everything in behalf of the user. \\
4. Restart the task from the beginning. Your environment state has been reset. \\
5. You are currently at the start of the task, but the provided previous trajectory covers the entire multi-turn interaction. Do NOT execute or simulate future steps ahead of time. Focus ONLY on the current task instruction. \\
6. You can use related memory to guide your reasoning to solve the problem. \\
\bottomrule
\end{tabular}
\end{table*}

\begin{table*}[t]
\centering
\small
\caption{RefCon trajectory generation prompt for SWE-bench Verified (Mini).}
\label{tab:swe-prompt}
\begin{tabular}{p{0.95\linewidth}}
\toprule
\textbf{Critique Trajectory Prompt for Self-Refine} \\
\midrule
You are an expert reviewer analyzing an AI assistant's multi-turn software engineering trajectory. Your job is to identify mistakes, missed files, and suboptimal debugging decisions. For each turn in the trajectory, evaluate: \\
1. Did the assistant explore the repository effectively? Did it locate the relevant source files and classes, or did it waste turns on unrelated directories? \\
2. Was the bug localization accurate? Check if the assistant correctly identified the root cause of the issue before attempting a fix. \\
3. Did the assistant use the environment and test tools correctly? For example, if a reproduction script was created, did the assistant analyze the output to guide the patch? \\
4. Was the generated patch functional and minimal? Identify if the assistant introduced unnecessary changes or failed to follow the repository's coding style. \\
5. Did the assistant follow a logical debugging sequence? (e.g., search $\rightarrow$ reproduce $\rightarrow$ fix $\rightarrow$ verify via tests) \\
Be specific about which turns have issues and what should be done differently. If a turn looks correct, briefly note it as OK. Focus most of your analysis on turns that seem problematic. \\
IMPORTANT: You must respond ONLY with your critique. Do not attempt to solve the task yourself. \\
\midrule
\textbf{Self-Refine Prompt for SWE-bench Verified (Mini)} \\
\midrule
You are an AI assistant tasked with resolving a GitHub issue in a complex repository. A reviewer has analyzed your previous attempt and provided a critique. \\
1. Read the critique carefully to understand the logic gaps or coding errors made in the previous trajectory. \\
2. Trust the Codebase: The critique is a helpful guide, but the current file content and test execution results are the absolute truth. If the critique suggests a fix that contradicts the actual code logic, prioritize the codebase. \\
3. Do not ask the user for any additional information; you are expected to navigate the repository and solve the issue autonomously. \\
4. Restart the task from the beginning. The repository and environment state have been reset to the original buggy state. \\
5. You are currently at the start of the task, but the provided previous trajectory covers the entire multi-turn interaction. Do NOT skip steps or assume the fix is already applied. Focus ONLY on the current task instruction. \\
6. You can use related memory to guide your reasoning to solve the problem. \\
\bottomrule
\end{tabular}
\end{table*}

\end{document}